\documentclass[letterpaper]{article}
\usepackage{aaai}
\usepackage{times}
\usepackage{helvet}
\usepackage{courier}
\usepackage{graphicx}
\usepackage{xcolor}
\usepackage{multirow}
\usepackage{mycomments3}
\usepackage{hyperref}

\begin{document}
%
\title{Increasing Fairness in Predictions Using Bias Parity Score Based Loss Function Regularization }
\author{}
\author{Bhanu Jain,\qquad Manfred Huber\qquad Ramez Elmasri\\
\\bhanu.jain@mavs.uta.edu\qquad huber@cse.uta.edu\qquad elmasri@cse.uta.edu\\}



\maketitle
\begin{abstract}
\begin{quote}
\mdel{An i}\madd{I}ncreasing utilization of machine learning based decision support systems emphasizes the need for resulting predictions to be both accurate and fair to all stakeholders. In this work we present a novel approach to increase a \madd{Neural Network} model's fairness during \mdel{the} training\mdel{ phase of neural networks}. We introduce a family of fairness enhancing regularization components that we use in conjunction with the traditional binary-cross-entropy based accuracy loss. These loss functions are based on Bias Parity Score (BPS), a score that helps quantify bias in the models with a single number. In the current work we investigate the behavior and effect of these regularization components on bias\mdel{ in the machine learning based decisions}. We deploy the\madd{m}\mdel{se loss functions} in the context of a recidivism prediction task \mdel{on two commonly used datasets} as well as on a census-based adult income dataset\mdel{ for predicting income}. The results demonstrate that with a good choice of fairness loss function we can reduce the trained model's bias without deteriorating accuracy even in unbalanced datasets. 
\end{quote}
\end{abstract}
\section{Introduction}
The use of automated decision support and decision-making systems (ADM) \cite{hardt2016equality} in applications with direct impact on people's lives has increasingly become a fact of life, e,g. in criminal justice \cite{kleinberg2016inherent,jain2020reducing,dressel2018accuracy}, medical diagnosis \cite{kleinberg2016inherent,ahsen2019algorithmic}, insurance \cite{baudry2019machine}, credit card fraud detection \cite{dal2014learned}, electronic health record data \cite{gianfrancesco2018potential}, credit scoring \cite{huang2007credit} and many more diverse domains. This, in turn, has lead to an urgent need for study and scrutiny of the bias-magnifying effects of machine learning and Artificial Intelligence algorithms and thus their potential to introduce and emphasize social inequalities  and systematic discrimination in our society. Appropriately, much research is being done currently to mitigate bias in AI-based decision support systems \cite{ahsen2019algorithmic,kleinberg2016inherent,noriega2019active,feldman2015computational,oneto2020general,zemel2013learning}.

\textbf{Bias in Decision Support Systems}.  As our increasingly digitiz\madd{ed}\mdel{ing} world \madd{generates and collects}\mdel{stows away} more data\mdel{ with the passing of each day}, \mdel{more and more} decision makers are \madd{increasingly} using AI based decision support systems. With this, the need to keep the decisions of these systems fair for people of diverse backgrounds becomes essential. Groups of interest are often characterized by sensitive attributes such as race, gender, affluence level, weight, and age to name a few. While machine learning based decision support systems often do not consider these attributes explicitly, biases in the data sets, coupled with the used performance measures can nevertheless lead to significant discrepancies in the system's decisions. For example, \madd{as} many minorities have traditionally not participated in many domains such as loans, education, employment in high paying jobs, receipt of health care\madd{, resulting datasets are often highly unbalanced.}\mdel{. This can lead to unbalanced datasets as the minority-based data may be combined with the majority sensitive attribute-based data.} Similarly, some domains like homeland security, refugee status determination, incarceration, parole, loan repayment etc., may be already riddled with bias against certain subpopulations\mdel{, even in the absence of AI based decision support system}. Thus human bias seeps into the datasets used for AI based prediction systems which, in turn, amplify it further. Therefore, as we begin to use AI based decision support system, it becomes important to ensure fairness for all who are affected by these decisions.

\textbf{Contributions}. We propose a technique that uses Bias Parity Score (BPS) measures to characterize fairness and develop\mdel{s} a family of corresponding loss functions that are used as regularizers during training of Neural Networks to enhance fairness of the trained models. The goal here is to permit the system to actively pursue fair solutions during training while maintaining as high a performance on the task as possible.
We apply the approach in the context of \madd{several}\mdel{six} fairness measures \mdel{based on parity across groups} and investigate multiple loss function formulations and regularization weights in order to study the performance as well as potential drawbacks and deployment considerations. In these experiments we show that, if used with appropriate settings, the technique \mdel{reduces bias in the models and outdoes the cross-entropy alone loss function mode on two recidivism related datasets as well as on a commonly used Adult Income dataset in terms of accuracy, false positive rate, false negative rate, true positive rate, and true negative rate. By using these \mdel{L}\madd{l}oss functions, we were able to} measurably reduce\madd{s} \madd{race-based} bias in \mdel{the} recidivism \madd{prediction,} \mdel{results for the two race-based cohorts} and demonstrate on the gender-based Adult Income dataset that the proposed method\mdel{, for appropriate choice of regularization loss function,} can outperform state-of-the art techniques aimed at more targeted aspects of bias and fairness. In addition, we investigate potential divergence and stability issues that can arise when using these fairness loss functions, in particular when shifting significant weight from accuracy to fairness.

\mdel{In the work presented here, we utilize a quantitative measure of parity as our concept of fairness in the form of Bias Parity Score applied to metrics that include  FPR, FNR, TPR, TNR, FPR and STP, as well as combinations thereof.}


\section{Notation} 
\label{se:notation}
In this paper we \mdel{will} present the proposed approach \mdel{to actively improve fairness in \mdel{the context of trained} Neural Network models} largely in the context of class prediction tasks with \madd{subpopulations}\mdel{groups of interest} indicated by a binary sensitive attribute. However, the approach can easily be extended to multi-class sensitive attributes \mdel{across which fairness is to be enforced} as well as to regression tasks\mdel{, both of which will be briefly discussed}. For the description of the approach in \mdel{the context of} our primary \mdel{prediction} domain, we use the following notation to describe the classification problem and measures to capture group performance measures \mdel{to be used} to assess prediction bias\mdel{:}\madd{.}

\mdel{For e}\madd{E}ach element of the dataset, $D=\{(X, A, Y)_i\}$, \mdel{each data point} is represented \madd{by}\mdel{as} an attribute vector $(X, A)$, where  $X \in R^d$ represents \madd{its}\mdel{the \mdel{quantified}} features \mdel{of each element}\mdel{ in the dataset,} and $A \in \{0, 1\}$ is the \mdel{binary} sensitive attribute indicating the group \madd{it}\mdel{the data item} belongs to. $C\:= c(X, A) \in \{0, 1\}$ indicates the value of the  predicted variable, and $Y \in \{0, 1\}$ is the true value of the target variable\mdel{ in the training set}.

To obtain fairness characteristics, we use parity over statistical performance measures, $m$, across the two groups indicated by the sensitive attribute, where $m(A\madd{=a})$ indicates the measure for the subpopulation where \mdel{the sensitive attribute} $A=\mdel{1}\madd{a}$ \mdel{, and $m(A')$ representing the measure for the subpopulation where $A=0$}.

Since this paper will introduce the proposed framework in the context of binary classification\mdel{ prediction}, the fairness related performance measures used \mdel{here} will center around \mdel{different} elements of the confusion matrix\madd{, namely}\mdel{ and in particular} false positive\mdel{ rates}, false negative\mdel{ rates}, true positive\mdel{ rates}, and true negative rates. In the notation used here, these metrics \madd{and}\mdel{as well as} the corresponding measures \madd{are}\mdel{can be represented as}:

\begin{description}

\item[False positive rate \madd{(FPR)}:] {$P(C = 1 ~|~ Y = 0)$. \\
$m_{FPR}( A=\mdel{0}\madd{a} ) = P(C = 1 ~|~ Y = 0, A=\mdel{0}\madd{a})$ \mdel{ \\ $m_{FPR}( A=1)= P(C = 1 ~|~ Y = 0, A=1)$.}}  
\item[False negative rate \madd{(FNR)}:] {$P(C = 0 ~|~ Y = 1).$ \\
$m_{FNR}( A=\mdel{0}\madd{a} ) = P(C = 0 ~|~ Y = 1, A=\mdel{0}\madd{a})$\mdel{   \\ $m_{FNR}( A=1)= P(C = 0 ~|~ Y = 1, A=1).$} }

\item[True positive rate \madd{(TPR)}:] {$P(C = 1 ~|~ Y = 1)$. \\
$m_{TPR}( A=\mdel{0}\madd{a} ) = P(C = 1 ~|~ Y = 1, A=\mdel{0}\madd{a})$\mdel{  \\ $m_{TPR}( A=1)= P(C = 1 ~|~ Y = 1, A=1)$.} }

\item[True negative rate \madd{(TNR)}:] {$ P(C = 0 ~|~ Y = 0)$. \\ 
$m_{TNR}( A=\mdel{0}\madd{a})= P(C = 0 ~|~ Y = 0, A=\mdel{0}\madd{a})$\mdel{. \\ $m_{TNR}( A=1)= P(C = 0 ~|~ Y = 0, A=1)$.} }

\end{description}

Strict parity in \mdel{any of} these or similar measures \madd{implies that they are identical}\mdel{is achieved when the  measure\madd{s}} for both groups\mdel{ are identical}, i.e. if $m(A\madd{=0})=m(A\mdel{'}\madd{=1})$.

\section{Related Work}\label{RelatedWork}

Some recent works \cite{ntoutsi2020bias,krasanakis2018adaptive} summarize bias mitigation techniques into three categories: i) preprocessing \madd{input data approaches} \cite{calders2009building,feldman2015certifying,feldman2015computational}\mdel{ input data approaches}, ii) in-processing approaches or training under fairness constraint that focuses on the algorithm \madd{\cite{celis2019classification,zafar2017fairness,iosifidis2020fabboo}} and, iii) postprocessing approaches that seek to improve the model \madd{\cite{hardt2016equality,mishler2021fairness}}. \add{\mdel{Another work}\madd{Expanding on this,} \cite{orphanou2021mitigating} sums up mitigating algorithmic bias \madd{as}\mdel{to be comprised of three steps:} detection of bias, fairness management, and explainability \mdel{M}\madd{m}anagement\madd{.} \mdel{where}\madd{Their} "fairness management" \madd{step} is the same as \madd{the} aforementioned "bias mitigation techniques"  but \madd{they} added "Auditing" (Fairness formalization) as an additional category\mdel{ to bias mitigation techniques}.}

The first \madd{bias mitigation} technique involving preprocessing \mdel{input} data is built on the premise that disparate impact in \madd{training} data results in disparate impact in the classifier\mdel{ trained on such data}. Therefore\mdel{,} these techniques are comprised of massaging data labels and reweighting tuples\mdel{ of the dataset}. Massaging is altering the class labels that are deemed to be mislabeled \madd{due to}\mdel{because of} bias, while reweighting \mdel{involves} increas\madd{es}\mdel{ing} weights of some \mdel{of the} tuples over \mdel{the} others in the dataset. Calders et. al. \cite{calders2009building} assert that these massaging and reweighting techniques yield a classifier that is less biased than without such a process. They also note that while \mdel{the} massaging labels \mdel{method} is intrusive \mdel{in nature} and can have legal implications \cite{barocas2016big}, \mdel{the} reweighting \mdel{method on the dataset to make the labels not dependent on the sensitive attribute} does not have these drawbacks. 

The second technique of \madd{in-processing} fairness under constraint \add{\cite{celis2019classification,zafar2017fairness}} chooses \mdel{one or more of the disparate} impact metrics \cite{iosifidis2020fabboo,zafar2017fairness,calders2013controlling}\mdel{. This is}\madd{,} followed by modifying the imposed constraints during \mdel{the} classifier training or by addi\madd{ng}\mdel{tional} \mdel{linear program} constraints that steer the model towards the optimization goals \cite{zafar2017fairness,calders2013controlling,iosifidis2020fabboo,oneto2020general}. \madd{Similarly,}\mdel{For example, in a recent work Iosifidis et. al.}  \cite{iosifidis2020fabboo} change the training data distribution and monitor the discriminatory behavior of the learner.  When this discriminatory behavior exceeds a threshold, they \mdel{facilitate different fairness notions by} adjust\mdel{ing} the decision boundary to prevent discriminatory learning.    

The third technique to improve the results involves a postprocessing approach to comply with the fairness constraints.  The approach can entail selecting a criterion for unfairness relative to a sensitive attribute while predicting some target. In the presence of the target and the sensitive attribute, \mdel{Hardt et. al.}  \cite{hardt2016equality}, show\madd{s} how to adjust the predictor to eliminate discrimination as per their definition.

\add{Our \mdel{current} work \mdel{is on}\madd{uses} supervised learning \mdel{and we do not}\madd{without} balanc\mdel{e}\madd{ing} the dataset\mdel{ for both training and test data}. \madd{The reason is to avoid issues from massaging labels and the consideration that while reweighting can be effective for simple sensitive attributes and predictions, in general multiple overlapping sensitive attributes and classes could exist, making reweighting \mdel{to balance the dataset} complex and potentially impossible.} Some \mdel{current} work\mdel{s on} \madd{employing} unsupervised \mdel{learning using} clustering \mdel{algorithms} also seek to maintain the original balance in the dataset\mdel{ and have clusters that \mdel{have}\madd{maintain the} same proportion of types as the overall population}  \cite{abbasi2021fair,abraham2019fairness,chierichetti2018fair}.  
}

\textbf{Definitions of Fairness}. \mdel{Current l}\madd{L}iterature \mdel{on fairness} \mdel{recommends}\madd{contains} several \mdel{formal} concepts of fairness \mdel{which} requir\madd{ing}\mdel{e that} one or more demographic or statistical properties \madd{to be}\mdel{are} \mdel{held}\madd{constant} across \mdel{multiple} subpopulations\mdel{ in the corpus}. Demographic parity\mdel{, also referred to as} \madd{(or} statistical parity\madd{)}, mandates that \mdel{the} decision rates are independent of the \mdel{values of a} sensitive attribute\mdel{ that represents membership of different subgroups.} \cite{noriega2019active,louizos2015variational,calders2009building,zafar2015learning} \mdel{(Calders, Kamiran, and Pechenizkiy ; Zafar et al. 2015; Louizos et al. 2015)}. For binary classification \mdel{problems} this \madd{implies}\mdel{is often mathematically represented as} $P(C = 1|A = 0) = P(C = 1|A = 1)$,  where $C \in \{0, 1\}$ is the \madd{system's} decision\mdel{ made by the system}. This\mdel{, criterion}, however, makes an \mdel{underlying} equality assumption between \mdel{the} subpopulations which might not hold\mdel{ for all problems}. To address this,  
\mdel{several recent works} \cite{hardt2016equality} focus on error rate balance where fairness requires subpopulations to have equal \mdel{false positive rates (}FPR\mdel{) or equal false negative rates (}\madd{, } FNR\mdel{)}\madd{, } or both. Another common\mdel{ly used} \mdel{parity} condition is equality of odds \madd{demanding}\mdel{which commands} equal \mdel{true positive rate (}TPR\mdel{)} and \mdel{equal true negative rate (}TNR\mdel{)}. While perfect parity \mdel{as a constraint} would be desirable, it often is not achievable and thus \mdel{quantitative} measures \madd{of}\mdel{representing} the degree of parity have to be used
\cite{jain2020using}. 
Refer to \cite {mehrabi2019survey,chouldechova2018frontiers,verma2018fairness} for a more complete \mdel{recent} survey of computational fairness metrics.

\madd{Here} \mdel{In the work presented here,} we \madd{use}\mdel{utilize} a quantitative \madd{parity} measure \mdel{of parity} as our concept of fairness in the form of Bias Parity Score \madd{(BPS)} applied to metrics that include  FPR, FNR, TPR, TNR, and FPR\mdel{, as well as combinations thereof}. 



\section{Approach}
\label{se:approach}

\mdel{As indicated above, achieving increased fairness or reduced bias in decision support systems is an important property and a range of techniques have been proposed to measure fairness.} T\madd{his paper}\mdel{he work presented here} is aimed at providing a framework to allow a deep learning-based prediction system to learn fairer models for known sensitive attributes by actively pursuing improved fairness during training. For this, we first need a quantitative measure of fairness that can be widely addressed. \mdel{For this w}\madd{W}e propose to use \mdel{the} Bias Parity Score (BPS) which evaluates the degree to which a common measure in the subpopulations described by the sensitive attribute is the same. Based on this fairness measure we then derive a family of corresponding \madd{differentiable} loss functions that \mdel{are differentiable and} can \mdel{thus} be used \mdel{as part of the training loss function} as a regularization term in addition to the original task performance loss.

\subsection{Bias Parity Score (BPS)}\label{se:BPS}
In machine learning based predictions, bias is the differential treatment of "similarly situated" individuals. It manifests itself in unfairly benefitting some groups\mdel{ while posing others at a\mdel{n unfair} disadvantage} \cite{jiang2020identifying,angwin2016machine,dressel2018accuracy,zeng2017interpretable,jain2019singular,jain2020reducing,ozkan2017predicting}. 
Bias in recidivism, for example, may be observed in a higher FPR \madd{and}\mdel{accompanied by} lower FNR for one subgroup\madd{, implying} \mdel{which implies}\mdel{ that \madd{its members}\mdel{individuals in this subgroups} are} more \mdel{often}\madd{frequent} incorrect\mdel{ly} predict\madd{ions}\mdel{ed} to reoffend and thus denied parole\mdel{, putting them at an unfair disadvantage}. 
Similar bias may be observed when predicting \madd{whom}\mdel{whether} to invite \mdel{job candidates} for interviews or \madd{when} making salary decisions, resulting in preferential \madd{treatment of} \mdel{hiring opportunities or salary offers to} one cohort\mdel{ and the opposite for another cohort}.

To capture this, the relevant property can be \madd{encoded}\mdel{captured} as a measure and parity between subgroups can be defined as equality for this measure. In this work we capture the similarity of the measures for the subgroups quantitatively in the form of Bias Parity Score (BPS).
\mdel{BPS is a fairness measure that helps us quantify the bias in a predictive model in a single number.} Given that $m_s(A=0)$ and $m_s(A=1)$ are the values of any given statistical measure, $s$, for the sensitive and non-sensitive subpopulations, respectively, BPS can be computed as: 
\begin{equation}
BPS_s = 100\frac{min(m_s( A=1 ), m_s( A=0 ))}{max(m_s(  A=1 ), m_s( A=0 ))} .
\end{equation}

where the multiplication factor of $100$ leads to a percentage measure, making it easier to read. 

A BPS of $100$ represents perfect parity \mdel{and thus the highest possible degree of fairness} while a BPS of $0$ represents maximal bias between the groups. \mdel{BPS of any statistical measure is thus the ratio of that statistical measure between two groups. 
The smaller and the greater statistical measure values are used in the numerator and the denominator, respectively, to compute a}\madd{This formulation yields a} symmetric measure for two groups\mdel{. BPS} \madd{and} provides us with a universal fairness measure that can be applied to any property across subgroups \mdel{and} to evaluate fairness in a\mdel{ny} predictive model generated using a\mdel{ny given} machine learning classifier. 

We here defined BPS in terms of a binary sensitive attribute, i.e. in the context of fairness between two groups. While this is the case we will investigate in this paper\mdel{ when evaluating the benefits of translating this to loss functions and using it to train a Neural Network predictor}, this fairness measure can \mdel{relatively} easily be expanded to \mdel{a situation with} a multi-valued sensitive attribute $A \in \{a_1, ... a_k\}$ in a form such as:
\[
BPS_s = \sum_{a_i} \frac{100}{k} \frac{min(m_s( A=a_i ), m_s( ))}{max(m_s(  A=a_i ), m_s( ))} 
\]

where $m_s( )$ is the value of the underlying measure for the entire population\madd{. BPS here} \mdel{and} thus measures fairness as the average bias of all classes compared to the population\mdel{ average. Similarly, the worst bias of any class co\madd{u}ld be used instead of the average}.

Even though, BPS can be used for many statistical measures
as described in \cite{jain2020using} 
we will use it \mdel{for evaluation purposes} here with \mdel{false-positive rate (}FPR\mdel{)}, \mdel{false-negative rate (}FNR\mdel{)}, \mdel{true-positive rate (}TPR\mdel{)}, \mdel{true-negative rate (}TNR\mdel{)}  as these are often used for classification systems. In addition we will \madd{employ}\mdel{utilize} it on prediction rate to obtain a BPS equivalent to quantitative statistical parity \mdel{in order} to facilitate comparisons with previous approaches in the Adult Income domain.

\subsection{BPS-Based Fairness Loss Functions}

While \madd{the} BPS score\mdel{s} of \madd{a} statistical entit\madd{y}\mdel{ies} \mdel{such as FPR} represent\madd{s} a measure of fairness, it does not lend itself directly to training a deep learning system since it is not generally differentiable. To address this, we need to translate the underlying measure into a differentiable form and combine it into a differentiable version of the  BPS score that can serve as a training loss function. As we are using FPR, FNR, TPR, and TNR here \mdel{as measures} we first have to define continuous versions of these functions. For this, we build our Neural Network classifier\madd{s} \madd{with}\mdel{to have} a logistic activation function as the output (or a softmax if using multi-attribute predictions), leading, when trained for accuracy using binary cross-entropy to the \madd{network} output, $y$, \mdel{of the network to} represent\madd{ing} the probability of the positive class, $y = P(Y=1 | X, A)$. \mdel{Note that i}\madd{I}n the experiments \mdel{performed,} we will withhold the sensitive attribute \mdel{from the input of the classifier} and thus \mdel{practically} $y=P(Y=1 | X)$\mdel{ in our experiments}. Using this continuous output, we can \mdel{re}define a continuous measure approximation $mc_s()$ for FPR, FNR, TPR, and TNR\mdel{ as}:
\begin{equation}
\begin{array}{l}
mc_{FPR}(A=k) =  \frac{\sum_{(X, A, Y)_i : A_i=k, Y_i=0} y_i}{\sum_{(X, A, Y)_i : A_i=k} y_i} \\
mc_{FNR}(A=k) =  \frac{\sum_{(X, A, Y)_i : A_i=k, Y_i=1} (1-y_i)}{\sum_{(X, A, Y)_i : A_i=k} (1-y_i)}\\
mc_{TPR}(A=k) =  \frac{\sum_{(X, A, Y)_i : A_i=k, Y_i=1} y_i}{\sum_{(X, A, Y)_i : A_i=k} y_i}\\
mc_{TNR}(A=k) =  \frac{\sum_{(X, A, Y)_i : A_i=k, Y_i=0} (1-y_i)}{\sum_{(X, A, Y)_i : A_i=k} (1-y_i)}\\
\end{array}
\end{equation}

\mdel{It is important to note that t}\madd{T}his is not equal to $m_s()$ as it is sensitive to deviations in the exact prediction probability\mdel{, f}\madd{. F}or example if the output changes from $0.6$ to $0.7$ the continuous measure, $mc_s()$, changes  while the full measure, $m_s()$, would not change since both \madd{values} would result in the positive class.

To reduce this discrepancy, we designed a second \mdel{measure} approximation, $ms_s()$, that uses a sigmoid function, $S(x)=\frac{1}{1+e^{-x}}$,  to more closely approximate the full statistical measure by reducing \mdel{the occurrences of} intermediate prediction probabilities:
\begin{equation}
\begin{array}{l}
ms_{FPR}(A=k) =  \frac{\sum_{(X, A, Y)_i : A_i=k, Y_i=0} S(y_i-0.5)}{\sum_{(X, A, Y)_i : A_i=k} S(y_i-0.5)} \\
ms_{FNR}(A=k) =  \frac{\sum_{(X, A, Y)_i : A_i=k, Y_i=1} S(0.5-y_i)}{\sum_{(X, A, Y)_i : A_i=k} S(0.5-y_i)} \\
ms_{TPR}(A=k) =  \frac{\sum_{(X, A, Y)_i : A_i=k, Y_i=1} S(y_i-0.5)}{\sum_{(X, A, Y)_i : A_i=k} S(y_i-0.5)} \\
ms_{TNR}(A=k) =  \frac{\sum_{(X, A, Y)_i : A_i=k, Y_i=0} S(0.5-y_i)}{\sum_{(X, A, Y)_i : A_i=k} S(0.5-y_i)} 
\end{array}
\end{equation}

\mdel{In the same way sigmoided version of the measures can be derived for the other statistics.}

Once measures are defined, a continuous approximation for BPS fairness for both \mdel{the continuous and the sigmoided} measures can be defined as\madd{:}
\begin{equation}
\begin{array}{l}
BPSc_s= \frac{min(mc_s( A=1 ), mc_s( A=0 ))}{max(mc_s(  A=1 ), mc_s( A=0 ))} \\
BPSs_s= \frac{min(ms_s( A=1 ), ms_s( A=0 ))}{max(ms_s(  A=1 ), ms_s( A=0 ))} 
\end{array}
\end{equation}

These, in turn can be \madd{inverted}\mdel{translated} into loss functions that can be used during training \mdel{by inverting it} and further expanded by allowing to weigh \mdel{the importance of} small \mdel{biases} versus large biases by raising the loss to the $k^{th}$ power which depresses the importance of fairness losses close to $0$ (i.e. when the system is almost \mdel{completely} fair).
\begin{equation}
\begin{array}{l}
LFc_{(s, k)}=\left(1-BPSc_s\right)^{k}\\
LFs_{(s, k)}=\left(1-BPSs_s\right)^{k}
\end{array}
\end{equation}

These loss functions are continuous and differentiable in all but one point, namely the point where numerator and denominator are equal and thus in the minimum of the loss function. This, however, can be easily addressed in the training algorithm when optimizing the overall loss function.

\subsection{Fairness Regularization for Neural Network Training}

The approach \mdel{proposed} in this paper is aimed at training Neural Network\mdel{-based} deep learning classifiers to obtain more fair results\mdel{ in decision support systems} \madd{while preserving prediction accuracy}. \mdel{The tasks we are addressing here are classification tasks with a goal of achieving high accuracy predictions that are also fair.} The underlying task is thus maximizing accuracy which is commonly encoded in terms of a binary cross entropy loss function, $LF_{BCE}$.

Starting from this, we utilize the fairness loss functions derived in the previous sections as regularization terms resulting in an overall loss function, $LFc$ and $LFs$ for continuous and sigmoided fairness losses, respectively:
\begin{equation}
\begin{array}{l}
LFc (\vec{\alpha}, \vec{k}) = LF_{BCE} ~+~ \sum_{s_i} \alpha_{i} LFc_{(s_i, k_1)}\\
LFs (\vec{\alpha}, \vec{k}) = LF_{BCE} ~+~ \sum_{s_i} \alpha_{i} LFs_{(s_i, k_1)}
\end{array}
 \end{equation}

where $\vec{\alpha}$ is a weight vector determining the contribution of each \mdel{of the different} fairness loss function\mdel{s to the fairness regularization}, $\vec{k}$ is a vector of powers to be used for each of the fairness losses, and $s$ is the vector of the \mdel{4} loss metrics, ${<FPR,FNR,TPR,TNR>}$. Setting an $\alpha_i$ to $0$ effectively removes the corresponding fairness criterion\mdel{ from the loss function}.

These loss function can be used to train a Neural Network classifier where different values for $\vec{\alpha}$, $\vec{k}$, and the choice of sigmoided vs continuous loss puts different emphasis on different aspects of the underlying fairness characteristics.

\begin{figure}[hb]
\centering
  \vspace*{-0.0in}\includegraphics[width=\columnwidth]{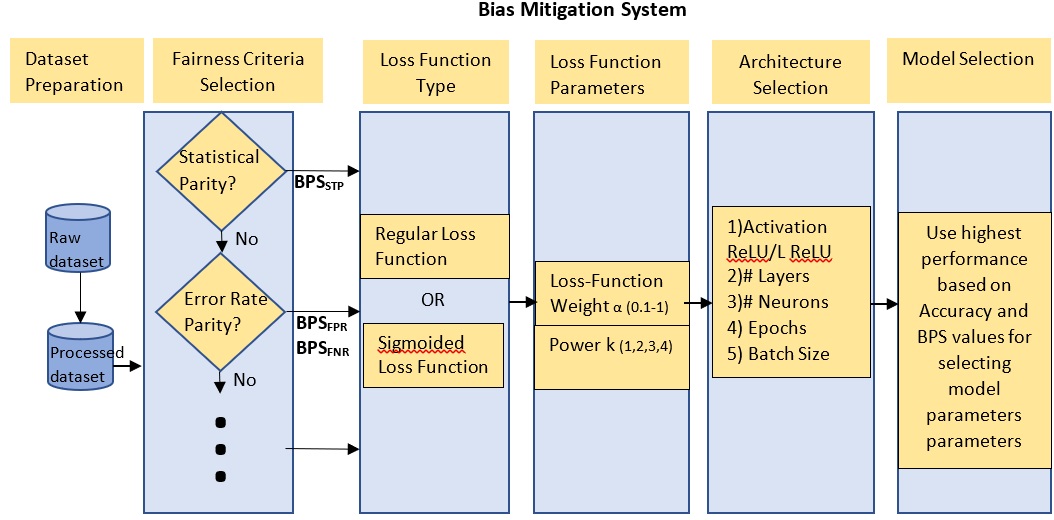}\\
  \vspace*{-0.025in}
  \caption{Design of a fair classifier\mdel{ in the proposed approach}. Based on selected fairness criteria, loss function type, \mdel{regression function} hyperparameter ranges, and network architecture, the best performing classifier is picked after a grid search based on the tasks's requirements.\vspace*{-0.125in}}
  \label{fi:diagram}
\end{figure}

\madd{Figure~\ref{fi:diagram} shows an overview of the basic model selection process for the proposed approach. Based on selected fairness criteria and hyperparameter ranges, a\madd{n}\mdel{ selected} architecture is trained in a grid search and the best model is selected.}

\section{Experiments}

To study the applicability of the proposed use of fairness losses as regularization terms, we conducted experiments on three different datasets, two in the the recidivism domain and one in \madd{the} income prediction domain, and analyzed the behavior and effects of different function and weight choices.

\vspace*{0.1in}\mdel{\subsection{Datasets}\label{Dataset}}
\textbf{Dataset~1:} \madd{This} is the main dataset used \mdel{in this study} and is the raw data from the study``Criminal Recidivism in a Large Cohort of Offenders Released from Prison in Florida, 2004-2008 (ICPSR 27781)'' \cite{NACJD27781website}\mdel{\madd{, available from}\mdel{. This dataset is available in} the ICPSR repository.~\footnote{\url{https://www.icpsr.umich.edu/icpsrweb/NACJD/studies/27781}} \mdel{and is based on the information provided by Florida Department of Corrections (FDOC) and the Florida Department of Law Enforcement (FDLE)}} \cite{bhati2014evaluating}. It \madd{contains}\mdel{is comprised of} 156,702 records \madd{with}\mdel{distributed in} a 41:59 \mdel{ratio of} recidivist\mdel{s} to non-recidivist\mdel{s} \madd{ratio}\mdel{records}. This ratio \mdel{of}\madd{for} our two \mdel{underlying} subpopulations is 34:66 for Caucasians and 46:54 for African Americans\madd{, making}\mdel{ which makes} it very unbalanced \mdel{between these groups} and lead\madd{ing}\mdel{s} to significant bias in traditional \mdel{recidivism prediction} approaches. In each crime category, the dataset has a higher proportion of non-recidivists Caucasians than African Americans. \mdel{The dataset}\madd{It} covers six crime categories and provides a large range of demographic features, including crime committed, age, time served, gender, etc.
We employed one-hot encoding \madd{for}\mdel{to treat} categorical features and trained \mdel{our system} to predict \mdel{whether an offender would be} reconvict\madd{ion}\mdel{ed} within \mdel{the next} 3 years. 

\textbf{Dataset~2:} \madd{This} is a secondary dataset \mdel{for this study} to validate results\mdel{ from the less balanced Dataset~1}.  This dataset ensued from \madd{the} ``Recidivism of Prisoners Released in 1994'' study \cite{NACJD03355website}. It contains data from 38,624 offenders that were released in 1994 from one of 15 states in the USA\madd{, with each record containing}\mdel{.  This is a very comprehensive dataset in the recidivism domain that contains} up to 99 pre and post 1994 criminal history records, treatments and courses taken by offenders\mdel{ while they were still in prison}. 
As described in 
\cite{jain2020reducing,jain2020using}
, each \mdel{of the 38,624} record\mdel{s} was split to create one record per arrest cycle\mdel{ and treat each arrest cycle as a point in time for a parole decision to be made}\mdel{. Any subsequent relapse \mdel{into criminal ways} was used to label records for recidivism}\madd{, resulting}\mdel{. This process resulted} in approximately 442,000 records that use demographic and history information to predict reconviction. This dataset is significantly more balanced\mdel{ between the two underlying subpopulations}, allowing \mdel{us} to verify effects \madd{from}\mdel{identified in} Dataset~1 in \mdel{a} data\mdel{set} with different characteristics.

\begin{figure*}[ht]
\centering
  \includegraphics[width=\textwidth,height=1.5in]{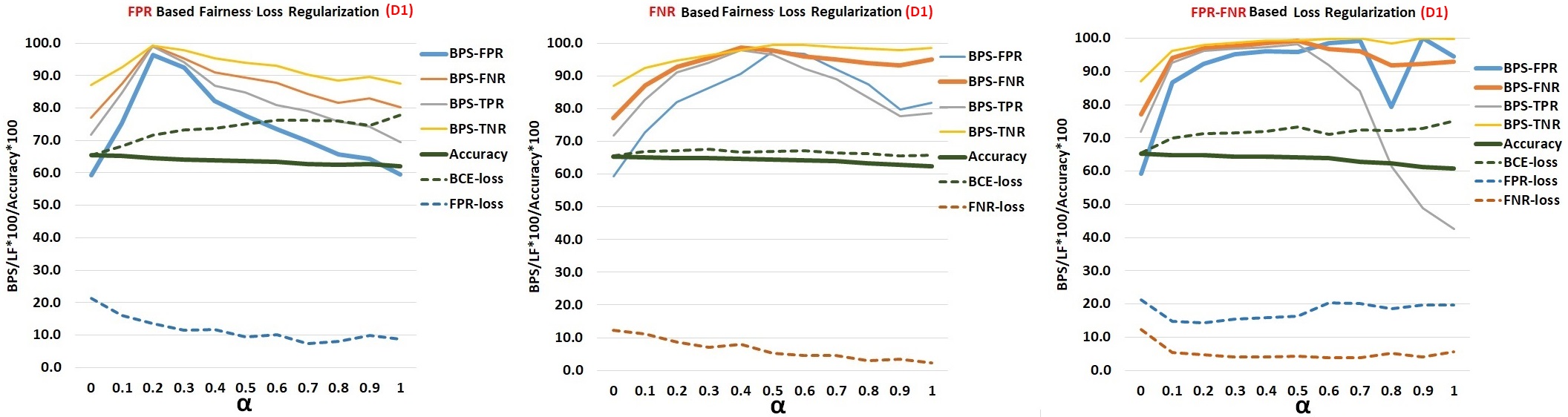}\\
  \vspace*{-0.15in}
  \caption{BPS Measures, Accuracy, and Loss function values as a function of regularization weight, $\alpha$, for  \madd{the continuous, linear case using} $LFc_{(FPR, 1)}$ \mdel{regularization} (left), $LFc_{(FNR, 1)}$ \mdel{regularization} (middle), and $LFc_{(FPR, 1)} + LFc_{(FNR, 1)}$ \mdel{regularization} (right).}
  \label{fi:linear}
\end{figure*}

\madd{\textbf{Dataset~3:} \madd{This} is the commonly used ``Adult Income Data Set'' \cite{kohavi1996scaling} from the UCI repository \cite{Dua:2019} that was extracted from 1994 Census data\mdel{base}. The dataset has 48,842 records\madd{,} \mdel{where} each \mdel{record} represent\madd{ing}\mdel{s} an individual \madd{over}\mdel{who was older than} 16\mdel{,} \madd{who} had an \mdel{adjusted gross} income greater than \$100\mdel{, worked for at least an hour a week and represented more than 1 person in the general population during the 1994 census}. The dataset holds socioeconomic status, education, and job information pertaining to the individual \madd{and is thus largely}\mdel{. Therefore, it is a} demographic\madd{,}\mdel{al dataset} just like Dataset 1\madd{,} but in a different domain\mdel{ than Datasets 1 and 2}. Unlike Datasets 1 and 2, where race is the sensitive attribute, gender is the sensitive attribute in Dataset 3 \madd{with the goal of predicting} \mdel{. For our work, similar to most previous work involving this data set, we deem} a salary higher than \$50 K\mdel{ a "positive" outcome and less than \$50 K to be a "negative" outcome. It is used for predicting whether an individual's income exceeds \$50K/yr or not and for related statistical measures}.
This dataset is \madd{here mainly} \mdel{also a demographic dataset and is} used to \madd{verify that effects translate to other domains and to \madd{permit}\mdel{perform} comparisons with \mdel{other} state-of-the-art techniques.}\mdel{evaluate bias mitigation techniques for gender-based inequities.}
}

\subsection{Constructing Neural Networks} 
For the \mdel{two} recidivism datasets \mdel{and all settings of the fairness regularization,} we trained \mdel{neural} networks with 2 hidden layers with 41 units \mdel{in} each\mdel{ of them}. Each \mdel{input and} hidden layer \mdel{was followed by}\madd{used} ReLU activation \mdel{function} and \madd{10\%} dropout \cite{srivastava2014dropout}\mdel{ with 10\% probability}\madd{, }\mdel{.  This was} followed by Batch Normalization\mdel{ of each layer. Batch Normalization}\mdel{ \madd{to} make\mdel{s} the optimization smoother} \cite{santurkar2018does}.  The output layer had 1 \madd{logistic} unit \mdel{which used a logistic output function} as indicated previously. We tuned various hyper parameters to select a batch size of 256 and 100 epochs and trained \mdel{our neural network model} using the Adam \cite{kingma2014adam} optimiz\madd{er}\mdel{ation algorithm for stochastic gradient descent}. \mdel{While training each parametrized model, we \madd{always} saved \mdel{it only if its}\madd{the highest} accuracy \mdel{improved over the one created in the previous epoch iteration}\madd{model}. }The \mdel{size, structure, and} hyperparameter\madd{s} \mdel{settings} \mdel{of the network} w\madd{ere}\mdel{as} chosen based on \madd{previous} experience \mdel{in the work presented in the previous chapters.
another work} with the\mdel{se} datasets. 
\cite{jain2020reducing,jain2020using,jain2019singular}.

\subsection{Evaluation Study}
To \mdel{assess the operation of and} evaluate the characteristics of \mdel{the proposed}
fairness loss regularization in \mdel{the context of the} recidivism \madd{prediction}\mdel{datasets}\mdel{ with race-based fairness criteria}, we conducted experiments with 6 \mdel{setting\madd{s} for the} measures \mdel{used in the fairness} for both continuous and sigmoided loss functions, employed 4 different exponents for the continuous case, and ran experiments for 10 different weight settings\mdel{ for the degree of influence of the fairness regularization}. In particular, we \mdel{chose settings that} used \mdel{each of} the 4 statistics individually as well as \mdel{ones that used} FPR and FNR or TPR and TNR simultaneously with equal weights. \madd{A grid search varied w}\mdel{W}eights $\alpha$ \mdel{were varied} between $0.1$ and $1$ in steps of $0.1$, and for \mdel{the} continuous loss \mdel{functions,} \madd{used} powers of $1$, $2$, $3$, and $4$\mdel{ used to train networks}. \mdel{The same experiments, except only using a power of 1 w\madd{ere}\mdel{as} repeated with sigmoided loss functions.} The goal \mdel{here} was to \mdel{be able to} compare the effects of different settings on the behavior of the system \mdel{both} in terms of accuracy\mdel{ achieved of the final model}, \mdel{the resulting} fairness, and \mdel{the} stability\mdel{ of the solution}. The Baseline \mdel{model} used \mdel{only BCE loss and} no fairness regularization.

To capture \mdel{the impact of} \mdel{training} variance\mdel{ on these models}, Monte Carlos cross validation with 10 iterations was \madd{used}\mdel{conducted} \mdel{for each of the setting} and the means of the \mdel{individual} metrics \mdel{for the resulting models} are reported\mdel{ \mdel{here} to increase reliability of \mdel{any} conclusions drawn\mdel{ and reduce the chance of outliers}}. \mdel{In addition, s}\madd{S}tability \mdel{of the resulting solutions} was evaluated using the variance\mdel{ across the 10 training runs}. \mdel{To study the effects, both BPS and loss \mdel{function} values were \mdel{recorded and} evaluated.}

\subsection{Results and Discussion}
The goal of th\madd{ese}\mdel{e set of} experiments \mdel{conducted here} is to evaluate whether the proposed approach \mdel{to active fairness training in deep learning systems through fairness regularization} can achieve the desired goal and to evaluate the effect of different loss function and regularization weight choices on the performance both in terms of accuracy and fairness. 
To perform this study \mdel{in the context of recidivism} we utilized mainly Dataset~1 due to its higher imbalance and studied the effect of different aspects of the loss function design\mdel{ in Sections~\ref{se:linweights}, \ref{se:sigmoid}, and \ref{se:power}}. We then used Dataset~2 to validate some of the results on a more balanced dataset\mdel{ in Section~\ref{se:balanced}}.

\begin{figure*}[ht]
\centering
  \includegraphics[width=\textwidth,height=1.5in]{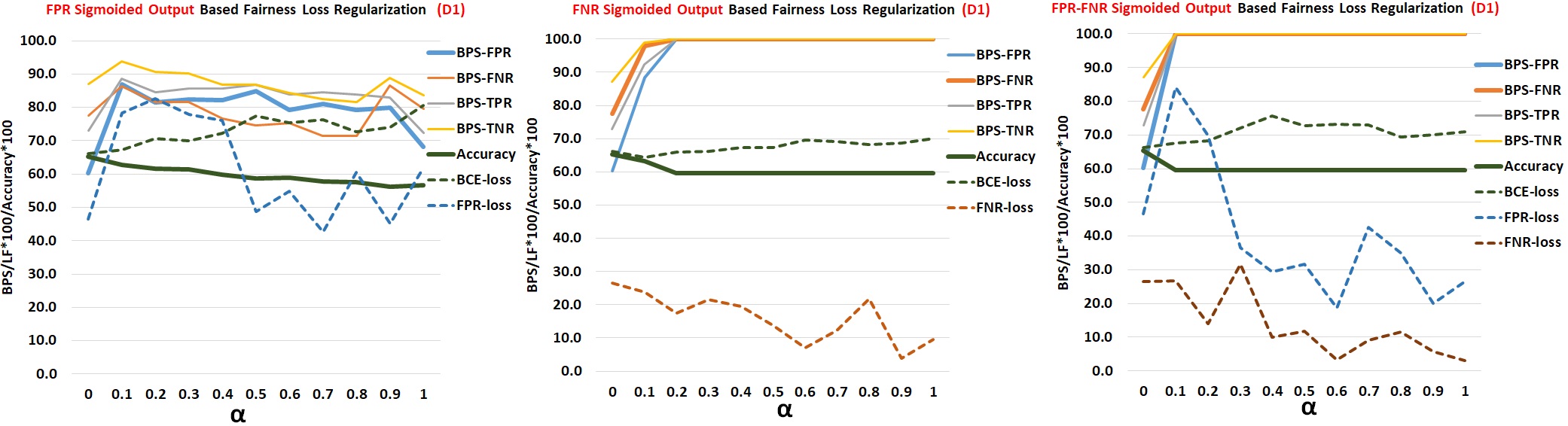}\\
  \vspace*{-0.15in}
  \caption{BPS Measures, Accuracy, and Loss function values as a function of regularization weight, $\alpha$, for  sigmoided loss functions $LFs_{(FPR, 1)}$ \mdel{regularization} (left), $LFs_{(FNR, 1)}$ \mdel{regularization} (middle), and $LFs_{(FPR, 1)} + LFs_{(FNR, 1)}$ \mdel{regularization} (right).}
  \label{fi:sigmoid}
\end{figure*}

\textbf{Measuring Bias in Results}: After conducting the experiments, we investigated the effect of the loss functions \mdel{by studying prediction results' characteristics for} \madd{on} residual bias \mdel{and} measured \mdel{these} by \mdel{computing the} Bias Parity Score for \mdel{the four main statistical measures,} FPR, FNR, TPR, and TNR, as well as \mdel{Bias Parity Score for}  Accuracy. In addition we recorded Accuracy as well as the values for BCE loss and for each of the fairness loss functions used in the respective experiment. \mdel{This allowed us to analyze effects of loss function choice and parameter settings on performance both in terms of the internal operation of the approach (i.e. loss functions) and of the desired performance metrics (Accuracy and BPS scores).}

\begin{figure*}[ht]
    \includegraphics[width=\textwidth,height=1.5in]{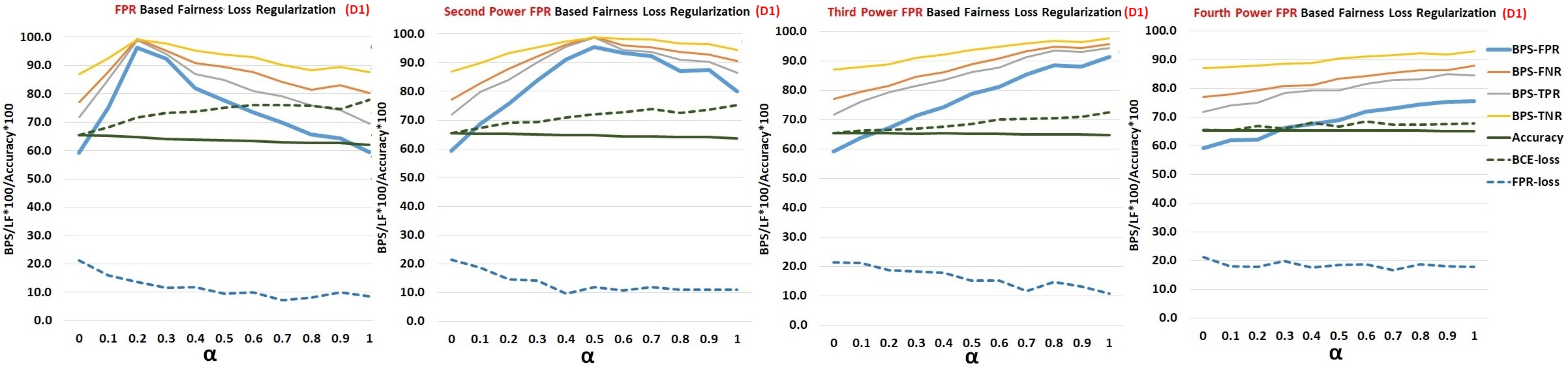}\\
  \vspace*{-0.3in}
    \caption{BPS Measures, Accuracy, and Loss function values as a function of regularization weight, $\alpha$, for  different powers of the continuous loss \mdel{function} functions $LFc_{(FPR, 1)}$ \mdel{regularization} (left), $LFc_{(FPR, 2)}$ \mdel{regularization} (center-left), $LFc_{(FPR, 3)}$ \mdel{regularization} (center-right), and $LFc_{(FPR, 4)}$ \mdel{regularization} (right).}
  \label{fi:powers}
\end{figure*}

\subsubsection{Applicability and Effect of Regularization Weight}\label{se:linweights}

To study the basic performance \mdel{of the system} \madd{we analyzed the experiments for} \mdel{as well as the effect on fairness and accuracy, a set of experiments were conducted for each of} the six continuous fairness measures in the linear case (i.e. with a power of 1). \madd{As indicated}\mdel{In these experiments}, the regularization weight was increased step-wise, starting from the Baseline with no regularization ($\alpha=0$) \mdel{and moving} to equal contributions of BCE \madd{and fairness}\mdel{loss and regularization} ($\alpha=1$). Figure~\ref{fi:linear} shows the average \madd{accuracy (solid grey line),} BPS scores \madd{(solid lines)}, \mdel{accuracy result,} and loss function values \madd{(corresponding dashed lines)} as a function of the regularization weight, $\alpha$. \mdel{Values for Accuracy and Loss \mdel{functions} are multiplied by $100$ to be in the same scale as BPS scores.} \mdel{In this figure,} \madd{Only} results for \madd{$FNR, FPR,$ and $FNR+FPR$-based regularization}\mdel{$LFc_{(FNR, 1)}$, $LFc_{(FPR, 1)}$, and $LFc_{(FNR, 1)} + LFc_{(FPR, 1)}$ experiments} are shown\mdel{ \madd{due to space limitations}}. Behavior for \madd{$TNR, TPR,$ and $TNR + TPR$}\mdel{$LFc_{(TNR, 1)}$, $LFc_{(TPR, 1)}$, and $LFc_{(TNR, 1)} + LFc_{(TPR, 1)}$} was similar.

\mdel{In all \mdel{of the} cases, t}\madd{T}hese graphs show that introducing \mdel{the} fairness regularization \mdel{loss} immediately \mdel{leads to an} increase\madd{s} \mdel{in} the corresponding BPS score, and thus \mdel{the} \mdel{corresponding} fairness\mdel{ measure}, while only \mdel{very} gradually decreasing accuracy\mdel{ of the prediction model as the influence of the fairness loss increases}. This demonstrates the viability of the \mdel{proposed} technique\mdel{ to obtain more fair prediction models for decision support systems without destroying task performance}. Moreover, in this case, regularization for one statistic also yields improvement in \mdel{the} other statistics\mdel{, at least initially}, which can be explained with the close relation of FPR, FNR, TPR, and TNR\mdel{ in the context of accuracy}\mdel{ as a performance metric}.

However, the experiments \mdel{with the three different loss functions} also show some \mdel{important} differences that give \mdel{important} information regarding important considerations \madd{for}\mdel{when determining} regularization weights. In particular, while the experiments with $LFc_{(FNR, 1)}$ and $LFc_{(FNR, 1)} + LFc_{(FPR, 1)}$ show a relatively steady increase in fairness\mdel{ as the regularization weight is increased}, the case of $LFc_{(FPR, 1)}$ shows that after an initial strong increase\mdel{ in fairness}, the \madd{active} fairness measure\mdel{ encoded in the regularization loss function}, $BPSc_{FPR}$, starts to decrease \mdel{and finally collapse} once \mdel{the regularization weight,} $\alpha$\mdel{,} exceeds $0.2$. At the same time \mdel{it can be observed that} the regularization loss, $LFc_{(FPR, 1)}$, continues to decrease, showing a decoupling between the core fairness measure and the loss function at this point. The reason \mdel{for this discrepancy} is that in order to obtain a differentiable loss function it was necessary to interpret the network output as a probabilistic prediction and thus\madd{, for example,} \mdel{the loss function can be improved \mdel{by changing the output $y_i$ from $0.4$ to $0.3$ which has no effect on}\madd{without changing the actual predictions and thus} the \mdel{actual} fairness measure.} \mdel{D}\madd{d}ecreasing the output for one negative item from $0.4$ to $0.1$ while simultaneously decreasing an output for a positive data item from $0.51$ to $0.49$ here represent an improvement in loss function value while reducing the corresponding fairness BPS as the second item classification became incorrect.

\subsubsection{Sigmoided Loss Function}\label{se:sigmoid}

One way to address th\madd{is}\mdel{e} decoupling \mdel{of the loss function from the fairness measure} \mdel{is the use of}\madd{are} the proposed sigmoided fairness loss function\madd{s}\mdel{ which aggressively moves metric values for the loss function towards $0$ and $1$}. Figure~\ref{fi:sigmoid} shows the \madd{corresponding} results \mdel{for the corresponding cases} \mdel{to the previous section} \mdel{but} using the sigmoided version of the fairness loss \mdel{function for} $LFs_{(FNR, 1)}$, $LFs_{(FPR, 1)}$, and $LFs_{(FNR, 1)} + LFs_{(FPR, 1)}$. Again\madd{,} BPS values, Accuracy, and loss function values are shown.

While these results\madd{,} \mdel{when} compared to the linear results in Figure~\ref{fi:linear}\madd{,} as expected show less \mdel{of a} decoupling between loss \mdel{function} and fairness\mdel{ (except between the baseline and the first result with regularization weight of 0.1)}, tend\mdel{s} to impose higher levels of fairness earlier, and maintain fairness more reliably\mdel{ throughout the full range of regularization weights}, the\madd{y} \mdel{results} also show a stronger degradation in accuracy and, when looking at \mdel{the} regularization loss \mdel{function} for higher weights \madd{and corresponding variances} also exhibit strong signs of instabilities\mdel{, reflected in significantly increased variances across the 10 runs used in these results}. This implies that while there are advantages in terms of how well \mdel{the} sigmoided loss function\madd{s} represent\mdel{s} fairness, optimizing the\madd{m}\mdel{se regularization functions} is significantly harder \mdel{for the algorithm due to much steeper gradients}, leading to less stable convergence. When choosing between these two options it is thus important to consider this tradeoff and to have ways to monitor gradient stability.

\subsubsection{Effect of Loss Function Power}\label{se:power}

Another way to modify the effect of \mdel{the} regularization losses is to increase the \madd{loss function} power\mdel{ of the loss function}. \madd{This} \mdel{Raising the power} \mdel{will} reduce\madd{s} the impact of small amounts of bias near a fair solution while increasing importance \mdel{of the fairness measure} if fairness losses are high. The goal \mdel{here} is \mdel{to reduce small changes near the solution and thus} to reduce the small scale adjustments \madd{near a solution} most responsible for \madd{decoupling}\mdel{divergence between fairness and corresponding loss function}. Figure~\ref{fi:powers} shows the effect of different powers for \mdel{the} continuous loss \mdel{function}  $LFc_{(FPR, k)}$ \mdel{for powers $k$ or 1, 2, 3, and 4.}\madd{, the case} \mdel{This was the example} \mdel{in the initial, linear experiments} where \mdel{a} strong decoupling \mdel{between loss function and fairness metric} occurred.

These graphs show that as the power \mdel{of the loss function} increases, \mdel{the} improvement in fairness becomes smoother and \madd{decoupling of loss and fairness occurs}\mdel{the loss function and fairness measure diverge} \mdel{less and} significantly later. However, \mdel{\madd{they}\mdel{the graphs} also show that} a \mdel{significantly} larger \mdel{regularization} weight is \madd{needed}\mdel{required} to optimize\mdel{ the function}\mdel{ with a higher power}\mdel{ loss function}, with \mdel{the} power 4 \mdel{experiment} not reaching the best fairness\mdel{ even at \mdel{a weight of} $\alpha=1$}. \mdel{While this is not a problem \madd{here}\mdel{ in this case as the fairness loss function is relatively low}, it}\madd{This} might be\mdel{come} a problem in datasets \madd{with larger} \mdel{where} fairness loss\mdel{ is inherently larger}\mdel{, thus over-emphasizing the effect of fairness}. The best performance \mdel{here} seems \mdel{to be \mdel{achieved}} with power 3.

\subsubsection{Effects in More Balanced Data}\label{se:balanced}

\begin{figure}[h]
     \includegraphics[keepaspectratio,height=1.5in]{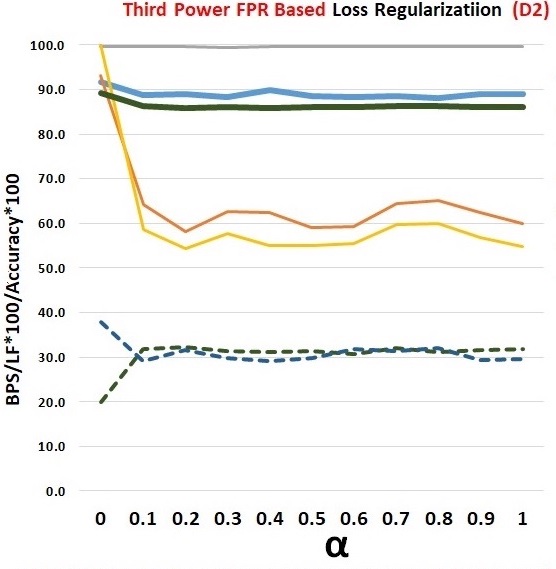}
     \includegraphics[keepaspectratio,height=1.5in]{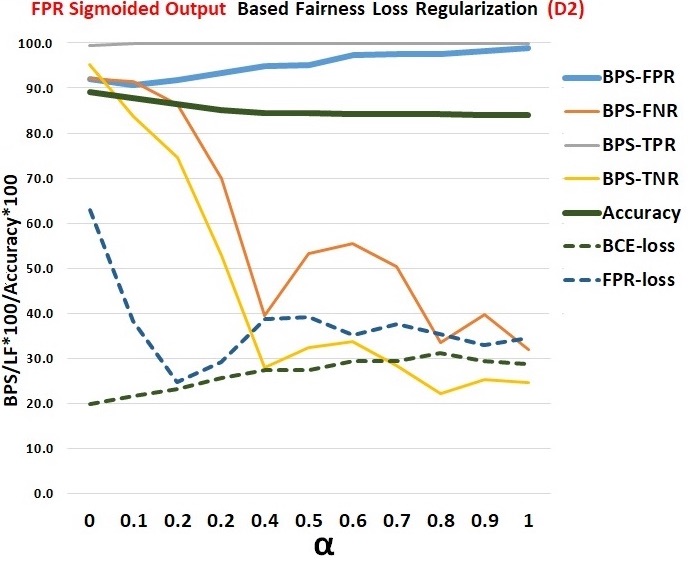}\\
  \vspace*{-0.3in}
           \caption{BPS\mdel{ Measures}, Accuracy, and Loss \mdel{function} values \mdel{as a function of regularization weight, $\alpha$,} for  \mdel{dataset} Dataset~2 with $LFc_{(FPR, 3)}$ \mdel{regularization} (\mdel{top}\madd{left}), and sigmoided $LFs_{(FPR, 1)}$ \mdel{regularization} (\mdel{bottom}\madd{right}).\vspace*{-0.0in}}
  \label{fi:D2}
\end{figure}

Dataset~1\mdel{,} \mdel{as described before, }contains relatively biased data, reflected in the \mdel{initially} \mdel{relatively} low \madd{base} fairness scores \mdel{in the range} of 70 - 80 \madd{and}\mdel{. Also, this dataset only contains relatively basic attributes, thus limiting achievable} accuracy \madd{of}\mdel{to} \mdel{around} 65\%. \madd{Thus it}\mdel{Due to this, this dataset} offers \mdel{relatively} significant room for improvement in fairness without \mdel{significant} \mdel{deterioration in}\madd{loss of} accuracy. To see if similar benefits can be achieved in \mdel{the context of a} less biased \mdel{and \mdel{significantly} richer} dataset\madd{s}, the experiments were repeated on \mdel{the} Dataset~2 \madd{which has}\mdel{. In this dataset} initial fairness \mdel{in the baseline case is} \madd{near}\mdel{closer to} 90 and accuracy \mdel{reach\madd{ing}}\mdel{es}\madd{of} 89\%. To see how similar \mdel{loss function} settings work \madd{here}\mdel{in this case}, Figure~\ref{fi:D2} shows \mdel{the} results for the best \madd{continuous and sigmoided} settings \mdel{\madd{on}\mdel{for} Dataset~2} \mdel{for both continuous and sigmoided loss function} for \mdel{the} FPR-based reguarization. In particular, it shows the continuous case with power 3, $LFs_{(FPR, 3)}$ and the \mdel{basic} sigmoided case of $LFs_{(FPR, 1)}$.
\begin{figure*}[ht]
  \includegraphics[width=\textwidth]{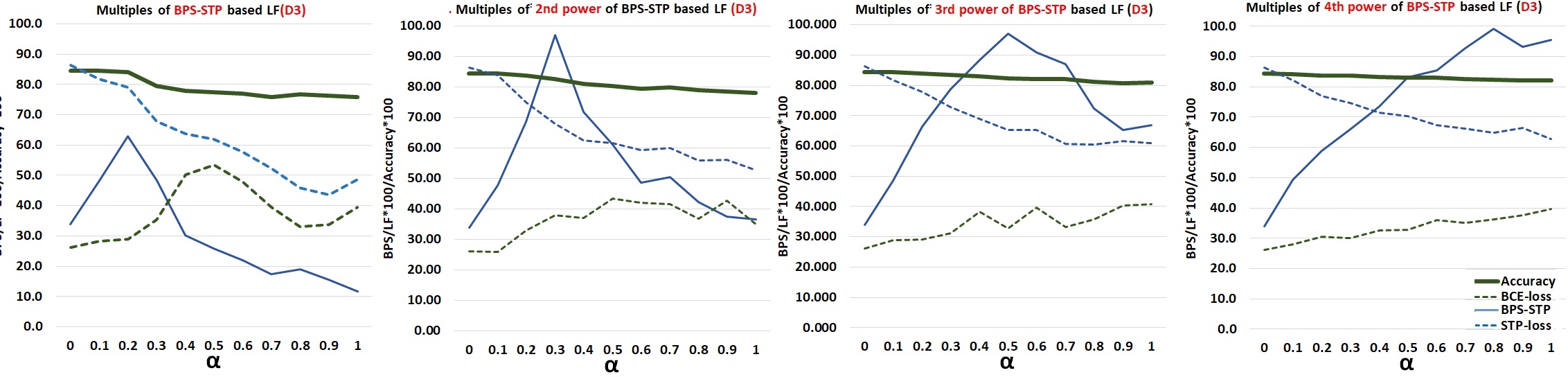}\\
  \vspace*{-0.3in}
  \caption{BPS Measures, Accuracy, and Loss function values as a function of regularization weight, $\alpha$, \madd{for Architecture 2 on Adult income data} for  $LFc_{(STP, 1)}$ \mdel{regularization} (\mdel{top-}left), $LFc_{(STP, 2)}$ \mdel{regularization} (\mdel{top-}right), $LFc_{(STP, 3)}$ \mdel{regularization} (\mdel{bottom-l}eft), and $LFc_{(STP, 4)}$ \mdel{regularization} (\mdel{(bottom-}right)\mdel{)} \mdel{Architecture 2:
  leaky ReLU Activation Fn. in the 2 hidden layers having 108 and 318 neurons respectively.} \vspace*{-0.15in}}
  \label{fi:Architecture2}
\end{figure*}

These results show that even for \mdel{the} less biased dataset\madd{s} where \mdel{improvement in} fairness \madd{gain} is more difficult\mdel{ to achieve}, the proposed \mdel{regularization} method \mdel{is successful at}\madd{can} increas\madd{e}\mdel{ing} \mdel{the desired} fairness \mdel{characteristic} without a dramatic drop in accuracy. However, \mdel{\madd{in contrast to Dataset 1 results,}}\mdel{while in the first dataset an increase in fairness in one metric tended to go hand in hand with increases in fairness in the other metrics even though they were not explicitly used as part of the training, this effect does not occur in the Dataset~2. In order to} improve\madd{d} fairness in one metric \mdel{in this less biased dataset}\mdel{, improvement in one fairness metric} \madd{here} yields degradations in others. This is not \mdel{entirely} unexpected \mdel{here} as the smaller \mdel{amount of} improvement potential is bound to require more tradeoffs. 

\section{Performance \madd{on Adult Income Data}\mdel{Evaluation using Dataset~3}}\label{se:D3Exp}
To \mdel{demonstrate}\madd{show} \mdel{that the} applicability \mdel{and results for the proposed fairness approach extend} beyond \mdel{the} recidivism \mdel{domain and race-based bias,} \madd{and}\mdel{as well as} to \mdel{be able to} compare \mdel{overall} performance with other state of the art approaches, we applied our approach on Dataset~3 and compared the results with those \mdel{published} in \cite{krasanakis2018adaptive,krasanakis2018adaptive,beutel2017data,zafar2017fairness,kamishima2012fairness}. 
\mdel{To do these comparisons we again evaluated effects of different settings in the loss functions but also experimented with multiple Neural Network architectures and certain data pre-processing methods, including data normalization.}

\subsection{Neural Network Architectures}

To \mdel{determine} \madd{find} the best network \mdel{architecture} we experimented with a number of hyperparameters\mdel{,} \mdel{varying the number of layers, the number of units in each layer,  the activation functions used in the hidden layers, and the drop out rate to find the architecture with the highest accuracy. \mdel{As a result of this, we developed}\madd{settling on} two architectures \madd{to test:}\mdel{that were used in the experiments.}} \madd{to find the network with the highest baseline accuracy\mdel{ without regularization}. Starting from Architecture~1 \mdel{the architecture used for} from the recidivism datasets (two ReLU hidden layers, each with108 neurons), we derived Architecture~2 (two leaky ReLU hidden layers with 108 and 324 neurons, respectively), yielding a small baseline accuracy increase from 84.5\% to 84.75\%.}

\mdel{\textbf{Architecture 1}:  T\madd{his}\mdel{he first architecture} was directly derived from the ones \mdel{used} for recidivism datasets and was comprised of two hidden layers, each with ReLU activation function and 108 neurons.

\textbf{Architecture 2}: T\madd{his}\mdel{he second architecture} was modified based on experiences with the Dataset~3 and was comprised of two hidden layers, each with leaky ReLU activation function but 108 and 324 neurons respectively.

The main considerations in the design of the second architecture was the achieved accuracy when trained without any fairness considerations, representing a slight improvement in that condition from the baseline accuracy of 84.5\% for Architecture 1 to a baseline accuracy of 84.747\%  for Architecture 2.}

\subsection{Results and Comparison}\label{se:D3results}

\subsubsection{Effect of Loss Function Parameters}

Varying loss function \madd{parameters}\mdel{weights, continuous vs sigmoided loss functions, and loss function powers in the context of the Dataset~3} \madd{again} yielded very similar observations \mdel{as \madd{fort eh recidivism data}}\mdel{in the case of the recidivism datasets} as shown \mdel{by the graphs in Figure~\ref{fi:Architecture1} for Architecture 1 and}\madd{in} Figure~\ref{fi:Architecture2} for Architecture 2 \madd{and various powers}.  
\meqdel{\begin{figure*}[h]
\includegraphics[width=\textwidth]{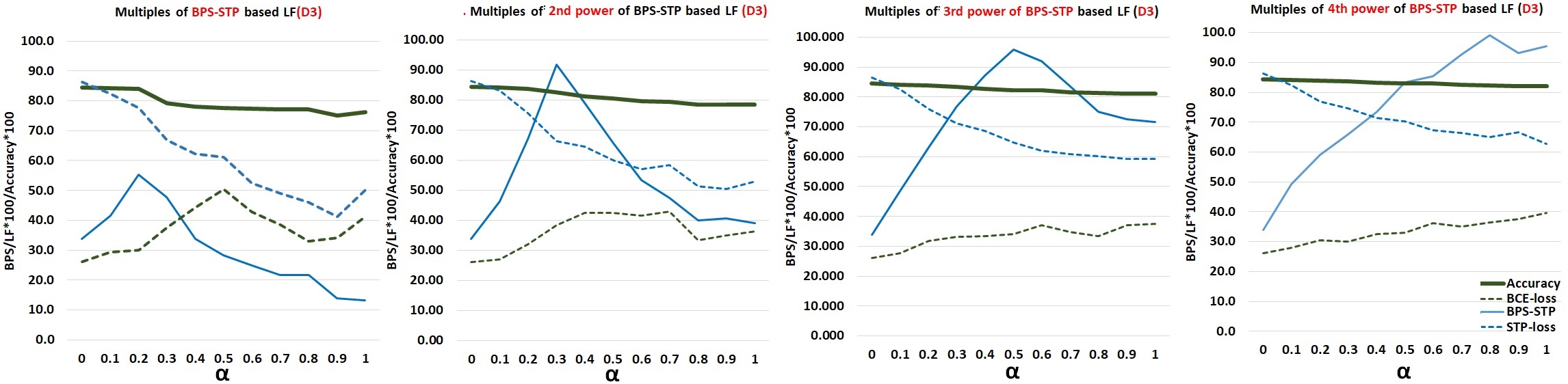}
  \caption{BPS Measures, Accuracy, and Loss function values as a function of regularization weight, $\alpha$, for  $LFc_{(STP, 1)}$ regularization (top-left), $LFc_{(STP, 2)}$ regularization (top-right), $LFc_{(STP, 3)}$ regularization (bottom-left), and $LFc_{(STP, 4)}$ regularization (bottom-right). Architecture 1.\\
  Architecture 1: ReLU Activation Fn. in the 2 hidden layers having 108 neurons each.}
  \label{fi:Architecture1}
\end{figure*}
}

The \mdel{figures}\madd{graphs} \mdel{here} show \mdel{the} results \mdel{for different powers (1, 2, 3, and 4)} \madd{for}\mdel{using} positivity rate  ($P(C=1 | X)$) as the underlying measure which, as a BPS score corresponds to \mdel{a BPS version of the} Statistical Parity\mdel{ fairness measure}. 
As \madd{previously,}\mdel{in previous experiments,} increasing the power reduced decoupling\mdel{ between the loss function and the BPS score and led to smoother improvements}. \mdel{Additional s}\madd{S}tudies with sigmoided loss \mdel{function} \mdel{on Dataset~3} further demonstrated the same behavior as for the recidivism data\mdel{sets}, demonstrating the \madd{framework's} generality\mdel{ of the proposed framework}. \mdel{In this case, however,}\madd{Here,} power 4 on continuous loss \mdel{functions seemed to} achieve\madd{d} the best results.

\mdel{\subsubsection{Effect of Architectures on D3 Dataset}.
We used two architectures for our experiments: Architecture 1 and Architecture 2. These were selected after tuning hyperparameters like number of neurons, number of epochs, weights associated with the loss function and the loss function itself. 
Architecture 1, was comprised of two hidden layers, each with ReLU activation function and 108 neurons. Architecture 2, was comprised of two hidden layers, each with leaky ReLU activation function but 108 and 324 neurons. Both architectures produced very similar patterns as were shown with Dataset~1 and Dataset~2 too. However, Architecture 2 improved BPS scores further as shown in \ref{fi:Architecture1} and \ref{fi:Architecture2}. In particular, Architecture 2 achieved slightly higher accuracy at the same level of BPS score and was thus chosen for comparison experiments.}

\subsubsection{Comparison with State of The Art Results}

To compare the approach \mdel{proposed here} to \madd{recent} previous work, we compared \madd{it}\mdel{to the recent results} to \madd{\cite{krasanakis2018adaptive}}\mdel{Zhang's work} that optimized p-rule (i.e. statistical parity) \mdel{\cite{zhang2018mitigating}}, and \madd{to}\mdel{the work by Krasanakis} \cite{zhang2018mitigating} that used adversarial training to achieve results with closely matching FPR and FNR across genders.

\mdel{As \cite{zhang2018mitigating} used}\madd{To compare to}  p-rule\mdel{ as the fairness criterion}, we utilized BPS on prediction rate \madd{as our fairness measure}\mdel{ to incorporate corresponding loss functions into our approach}.
As \mdel{observed}\madd{seen} in the previous experiments\mdel{ on recidivism}, applying BPS-based fairness \mdel{loss in Dataset~3} again changed the overall accuracy only to \madd{a} small degree\mdel{ with an imposition of a very high degree of fairness loss to compare against prior p-rule work \cite{zhang2018mitigating}}, yielding a drop from 84.5\% to 82.3\% \mdel{as shown in Table \ref{tab:table-FprFnr} }using Architecture 1\mdel{,} \madd{while}\mdel{but while BPS-Acc improved from 86.6 to 89 and} p\mdel{R}\madd{-r}ule \madd{improved}\mdel{changed} from 33.9\% to 99.128\%. \madd{Using}\mdel{
By further changing to} Architecture 2 \madd{with a power 4 continuous loss \mdel{function} and $\alpha = 0.84$, p-rule and} \mdel{the} accuracy \mdel{could be} further improved \mdel{as shown in Figure~\ref{fi:Architecture2}. In this case, Architecture 2 with a power of 4 seemed to perform the best with weights around a value of 0.8 and thus a finer scaled grid search was performed for $\alpha$ values between $0.825$ and $0.842$ 
to find the best results at $\alpha=0.84$ with a $BPS_{STP}$ (pRule) of}\madd{to} $99.9\%$ and \mdel{an accuracy of} $83\%$\madd{, as shown in Table \ref{tab:table-STP}.}
\begin{table} [t]
\centering
 \caption{Adult Income \mdel{dataset} \mdel{disparate impact elimination\mdel{ for BPS-FPR-FNR-based Loss Function}.}  Accuracy and pRule Comparison\mdel{ with published results of other techniques}.\\
\emph{Architecture~1:} STP-Loss Function, k=4, $\alpha$=0.8\\
\emph{Architecture 2:} STP-Loss Function, k=4, $\alpha$=0.84
}
\label{tab:table-STP}
\begin{tabular}{ |l|c| } 
\hline

 \multirow{2}*{
  \textbf{Fairness Technique} }  & Adult Income \hspace{0.1cm}    \\
 
     & pRule \hspace{0.1cm} acc  \\

\hline
\cite{krasanakis2018adaptive} (no fairness) & 27\% \hfill 85\% \\
\hline
\cite{krasanakis2018adaptive} & 100\%  \hfill 82\%\\
\hline
 \cite{zafar2017fairness} & 94\% \hfill 82\% \\
\hline
 \cite{kamishima2012fairness} & 85\% \hfill 83\% \\
\hline
Baseline (\add{current work}) & 34\% \hfill 85\% \\
\hline
Architecture1 (\add{current work}) & 99\% \hfill 82\% \\
\hline
\textbf{Architecture 2} (\add{current work}) & \textbf{100}\% \hfill \textbf{83}\% \\

\hline

    \end{tabular}
\end{table}    
\mdel{Thus} \madd{As shown, we achieved}\mdel{our approach offered} higher accuracy than \cite{krasanakis2018adaptive} while maintaining a pRule of approximately 100\%.  
  
  

\mdel{The comparison of our results with both Architecture 1 and Architecture 2 to previous work with pRule-based fairness is shown in Table~\ref{tab:table-STP}, illustrating that the proposed approach can outperform the more specialized approaches, including ones used specifically for the pRule.}

To compare to the work in \cite{zhang2018mitigating} who \mdel{are} aim\mdel{ed at} \madd{to} achiev\madd{e}\mdel{ing} similar FNR and FPR values for both genders, we utilized a combined \mdel{fairness loss function that employed} $BPS_{FNR}$ and $BPS_{FPR}$ \madd{fairness loss}\mdel{ together in the training phase}.
The FPR and FNR \mdel{for adult income dataset} using our technique as shown in Table \ref{tab:table-FprFnr} were approximately equal across \mdel{the two} gender\madd{s}\mdel{ based groups} at 0.0589 versus 0.0628 and at 0.4431 versus 0.5105, respectively\mdel{. $BPS_{FPR}$ and $BPS_{FNR}$ for our results was a 94 and 87 respectively} with Architecture 1\mdel{, where a BPS of 100 means no unfairness while a BPS of 0 means complete unfairness. With}\madd{while} Architecture 2 \mdel{, we could} further improve\madd{d} $BPS_{FPR}$ and $BPS_{FNR}$ while maintaining low absolute values of FPR and FNR of the two gender based cohorts and thus outperformed the $BPS_{FPR}$ and $BPS_{FNR}$ in \cite{zhang2018mitigating}\mdel{ as shown in Table~\ref{tab:table-FprFnr}}. 


\begin{table} [tbp]
\small
\centering
\caption{Adult Income \mdel{dataset: False Positive Rate (}FPR\mdel{)} and \mdel{False Negative Rate (}FNR\mdel{)} \madd{Equality Comparison}\mdel{for income bracket prediction}\mdel{ \mdel{for the two gender based groups,} with and without debiasing \madd{for different appraocjes}}. \mdel{\\
Architecture 1:ReLU Activation Fn. in the 2 hidden layers having 108 neurons each.\\
Architecture 2:leaky ReLU Activation Fn. in the 2 hidden layers having 108 and 318 neurons respectively.\\}
\\Arch~1\madd{: }\mdel{ values with} FPR-FNR-Sigmoided-LF,\mdel{POW}\madd{k}=4,$\alpha_1$=0.05, $\alpha_2$=0.05\madd{.}\mdel{\\}
Arch~2\madd{:} \mdel{values with} FPR-FNR-Sigmoided-LF,\mdel{POW}\madd{k}=3,$\alpha_1$=0.1, $\alpha_2$=0.125
}
\label{tab:table-FprFnr}
\begin{tabular}{ |l|c|c|c|c| } 

 \hline

 \multirow{2}*{ }
&     & Female \hspace{0.1cm}   & Male \hspace{0.1cm} & BPS \hspace{0 cm} \\
 
 \vspace{0.02cm}
&     & Without \hspace{0.1cm} With  & Without \hspace{0.1cm} With &  \hspace{0.1cm}\\

 \cline{3-5}

 \multirow{2}*{\begin{tabular}[l]{@{}l@{}}Beutel et\\al. 2017\end{tabular}
 }
& FPR    & 0.1875 \hspace{0.1cm} 0.0308  & 0.1200 \hspace{0.1cm} 0.1778 & 17.3\hspace{0 cm}\\
 \cline{3-5}
 \vspace{0.02cm}
& FNR    & 0.0651 \hspace{0.1cm} 0.0822  & 0.1828 \hspace{0.1cm} 0.1520 & 52.6\hspace{0 cm}\\
 \hline
 \cline{3-5}
 \vspace{0.02cm}
 \multirow{2}*
 {\begin{tabular}[l]{@{}l@{}}Zhang et\\al. 2018
 \end{tabular}}
 & FPR    & 0.0248 \hspace{0.1cm} 0.0647  & 0.0917 \hspace{0.1cm} 0.0701 & 92.0\hspace{0 cm}\\
 \cline{3-5}
 \vspace{0.02cm}
 & FNR    & 0.4492 \hspace{0.1cm} \textbf{0.4458}  & 0.3667 \hspace{0.1cm} \textbf{0.4349} & 97.5\hspace{0 cm}\\
 \hline
 \cline{3-5}
 \vspace{0.02cm}
 \multirow{2}*{Arch~1}
 & FPR    & 0.0319 \hspace{0.1cm} 0.0589  & 0.1203 \hspace{0.1cm} \textbf{0.0628} & 93.9\hspace{0 cm}\\
 
 \cline{3-5}
 \vspace{0.02cm}
 & FNR    & 0.4098 \hspace{0.1cm} \textbf{0.4431}  & 0.3739 \hspace{0.1cm} 0.5105 & 86.8\hspace{0 cm}\\
\hline
 \vspace{0.02cm}
 \multirow{2}* {\textbf{ Arch~2} } 
 & FPR    &0.0319 \hspace{0.1cm} 0.0610 & 0.1203 \hspace{0.1cm} 0.0708 & \textbf{93.3}\hspace{0 cm}\\
 
 \cline{3-5}
 \vspace{0.02cm}
 & FNR    & 0.4098 \hspace{0.1cm} 0.4785  & 0.3739 \hspace{0.1cm} 0.4862 & \textbf{98.4}\hspace{0 cm} \\
\hline
 
 
 
    \end{tabular}
\end{table}

\section{Conclusions}
In this work we \mdel{have} proposed \madd{Bias Parity Score-based fairness metrics and an approach to}\mdel{elements and considerations to impose fairness on Neural Networks during the training phase. In particular we proposed to} translate \mdel{Bias Parity Score-based fairness metrics}\madd{them} into corresponding loss functions \madd{to be}\mdel{that can then be} used as regularization terms during training \madd{of deep Neural Networks} to actively achieve improved fairness \mdel{and reduced bias} between subpopulations\mdel{ in the data}. For this, we introduced a family of \mdel{derived} fairness loss functions and conducted experiments on recidivism prediction \mdel{data} \mdel{where we}\madd{to} investigate\mdel{d} \madd{the applicability and behavior of the \mdel{regularization} approach for different hperparameters}\mdel{regularization weights and fairness loss function settings that are added to the task function which represents accuracy through a binary cross entropy loss function}. \mdel{In these experiments w}\madd{W}e demonstrated \madd{the generality of the approach and \mdel{investigated and} discussed considerations for loss function choices.} \mdel{how to use the loss functions to bring measurable improvement in equity to predictions and hence to the cohorts involved.} \mdel{In contrast to some \mdel{of the other} previous works, t}\madd{T}his work does not depend on changing input or output labels to make fair recommendations while \mdel{simultaneously} not forsaking accuracy. \mdel{By building our loss functions and yet dropping the sensitive attribute information from the input feature vector of the neural networks model, we ensure that the this work is guided by fairness by unawareness guidelines. }

\madd{Additional experiments performed on a common Adult Income dataset to compare the approach to more specialized state-of-the-art approaches\mdel{ in terms of their performance metrics} showed that the proposed regularization framework is highly flexible and, when provided with an appropriate BPS-based fairness measure can compete with and even outperform these methods.}

\mdel{This work illustrates that by concurrently using one or more BPS measure-based loss functions in concurrence with binary cross entropy can design automated decision support systems that can optimize for social objectives such as fairness. However, this work also shows that a correct pick of the regularization weight and the fairness loss function form can be essential to address convergence stability and to address specifics challenges with a dataset, such as different levels of bias in the data and varying levels of improvement potential due to limitations in available attributes or initial performance. This work also shows that besides a correct weight, choice of power of the loss function or the use fo a sigmoided loss introduces both benefits and challenges, where use of sigmoided losses, for example,  can decrease the loss more rapidly and avoid divergence between fairness metric and loss function,  but can also destabilize the convergence.}

\mdel{
\section{Future Directions} 
In this paper we presented a family of loss functions and showed potential benefits and detriments of different choices. In future work we would like to investigate methods that could automatically adjust fairness regularization terms according to properties of the dataset used.

\add{
\emph{Beyond Neural Networks}:} There are many machine learning algorithms suitable for numerous applications in different domains and there is much room in those to increase the fairness in the model itself. 

\add{
\emph{Disjoint Sensitive Attribute}:} Like most of the fairness literature, we assume that the race attributes are accurate. There is an assumption that there is a disjointedness in the race attribute, however demographic groups intersect. There needs to be research on such a racial intersection.

\add{
\emph{Probabilistic Sensitive attribute }: It would be interesting to use a probabilistic model for the protected attributes with out bias mitigation approach. Some other works like \cite{mehrotra2021mitigating,chen2019fairness} consider a probabilistic model for the protected attributes when curtailing bias with noisy information }

}

\bibliographystyle{aaai}
\bibliography{bhanu-bibliography}

\end{document}